\documentclass[letterpaper, 10 pt, conference]{ieeeconf}  %

\usepackage[utf8]{inputenc}

\IEEEoverridecommandlockouts                              %

\usepackage[american]{babel}
\usepackage{cite}
\usepackage{float}
\usepackage{microtype}
\usepackage{todonotes}
\usepackage{mathtools}
\usepackage{graphicx}
\usepackage{todonotes}
\usepackage{booktabs}
\usepackage{url}
\usepackage{subcaption}
\usepackage{amssymb}
\usepackage{amsmath}
\usepackage{hyperref}

\addto\extrasamerican{%

}
\usepackage{balance}

\title{\LARGE \bf
Towards 6D MCL for LiDARs in 3D TSDF Maps on Embedded Systems with GPUs
}

\author{Marc Eisoldt$^{1}$, Alexander Mock$^{2}$, Mario Porrmann$^{1}$ and Thomas Wiemann$^{3}$%
\thanks{$^{1}$ Computer Engineering Group, Osnabrück University, Osnabrück, Germany
        {\tt\small firstname.lastname@uni-osnabrueck.de}}%
\thanks{$^{2}$ Knowledge-based Systems Group, Osnabrück University, Osnabrück, Germany
	{\tt\small amock@uni-osnabrueck.de}}%
\thanks{$^{3}$ Robotics in Computer Science, Fulda University of Applied Sciences, Fulda, Germany and DFKI Niedersachsen, Plan-based Robot Control, Osnabrück, Germany
        {\tt\small thomas.wiemann@informatik.hs-fulda.de}}%
\thanks{The DFKI Niedersachsen Lab (DFKI NI) is sponsored by the Ministry of Science and Culture of Lower Saxony and the VolkswagenStiftung}
}

\begin{document}

\maketitle
\thispagestyle{empty}
\pagestyle{empty}

\begin{abstract}

Monte Carlo Localization is a widely used approach in the field of mobile robotics.
While this problem has been well studied in the 2D case, global localization in 3D maps with six degrees of freedom has so far been too computationally demanding.
Hence, no mobile robot system has yet been presented in literature that is able to solve it in real-time.
The computationally most intensive step is the evaluation of the sensor model, but it also offers high parallelization potential.
This work investigates the massive parallelization of the evaluation of particles in truncated signed distance fields for three-dimensional laser scanners on embedded GPUs.
The implementation on the GPU is 30 times as fast and more than 50 times more energy efficient compared to a CPU implementation.

\end{abstract}

\section{Introduction}

Localization in maps is one of the fundamental problems in mobile robotics. 
Many applications like navigation or exploration rely on a known pose of a robot in a given map.
Multimodal probabilistic methods are the state of the art for self-localization in GPS-denied, marker-less regions.
The exact calculation of such models is too expensive, so particle filters are used to approximate the problem. 
Monte Carlo Localization (MCL) uses particles to sub-sample the infinite state space. 
Global localization with 3D poses in 2D maps can be efficiently solved with CPUs, as the number of required particles is small.
In realistic applications using smart re-sampling, only a few thousand particles are required in 2D.

3D maps can be generated with LiDAR sensors in short time with high accuracy~\cite{eisoldt2022fully}.
LiDAR sensors are small and lightweight, allowing them to be used on drones~\cite{rahn2023redrose}.
To localize such vehicles with MCL similar to the 3D case, would allow to develop autonomous flying systems for GPS-denied environments without relying on artificial beacons, allowing new applications for instance in autonomous maintenance and logistics.
With this work, we to provide a MCL implementation for 3D maps and 6D poses.
The challenge is, that for 6D poses in 3D maps, the number of required samples increases by several orders of magnitude. 
Currently, this problem can not be solved in real-time on standard CPUs.
Particle filters offer great potential to benefit from hardware architectures that allow massively parallel computations. 
Accordingly, we aim to exploit such parallelism on embedded GPUs to achieve real-time localization.

Localization requires 3D map representations, that support quick simulation of hypothetical LiDAR data.
For that, Truncated Signed Distance Fields (TSDF) are apt candidates.
They allow to compute the required sensor update efficiently, i.e., measuring the scan-to-map divergence for a certain LiDAR measurement at a certain pose hypothesis, as the closest distances to the surface can be directly read from the TSDF.

\begin{figure}[t]
	\centering
	\begin{subfigure}[b]{.325\linewidth}
		\colorbox{white}{\includegraphics[trim={0cm 0cm 0cm 0cm},clip,angle=0,width=0.99\columnwidth]{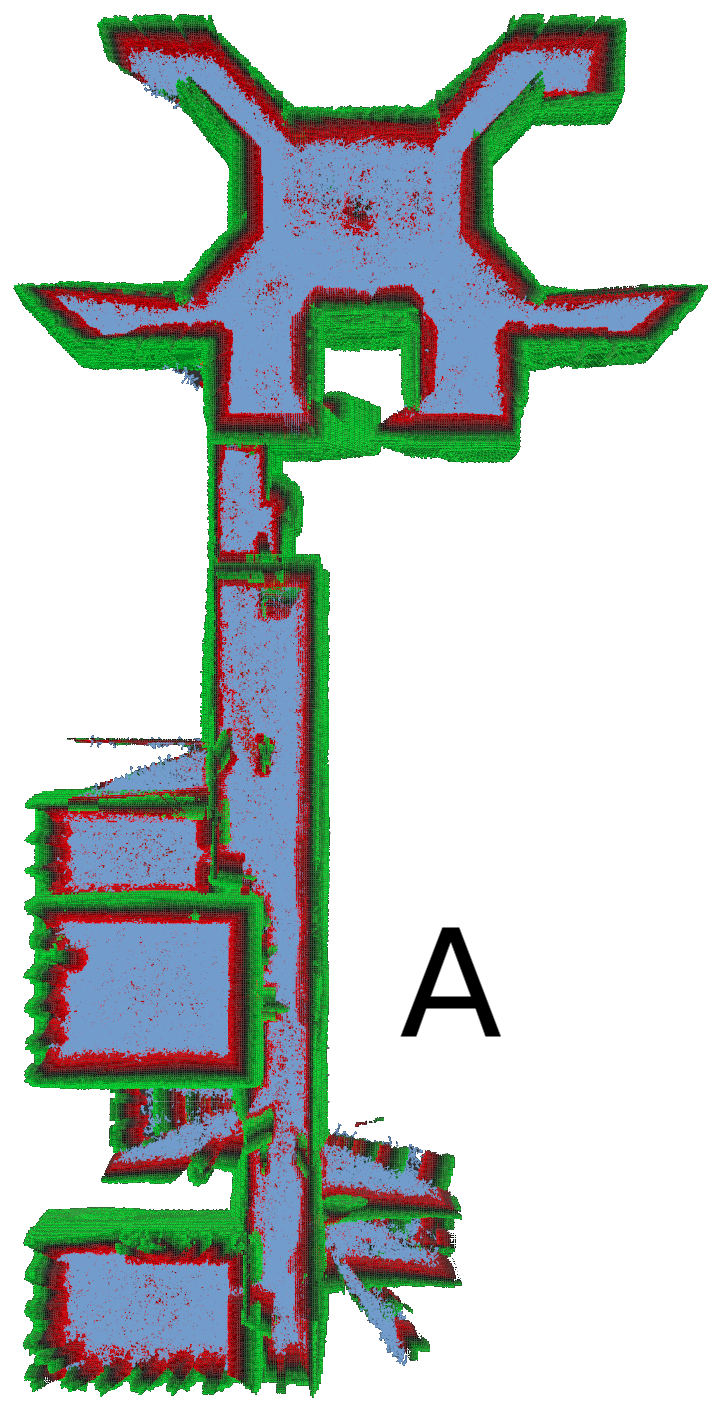}}
	\end{subfigure}
	\begin{subfigure}[b]{.32\linewidth}
		\includegraphics[trim={0cm 0cm 0cm 0cm},clip,angle=0,width=0.99\columnwidth]{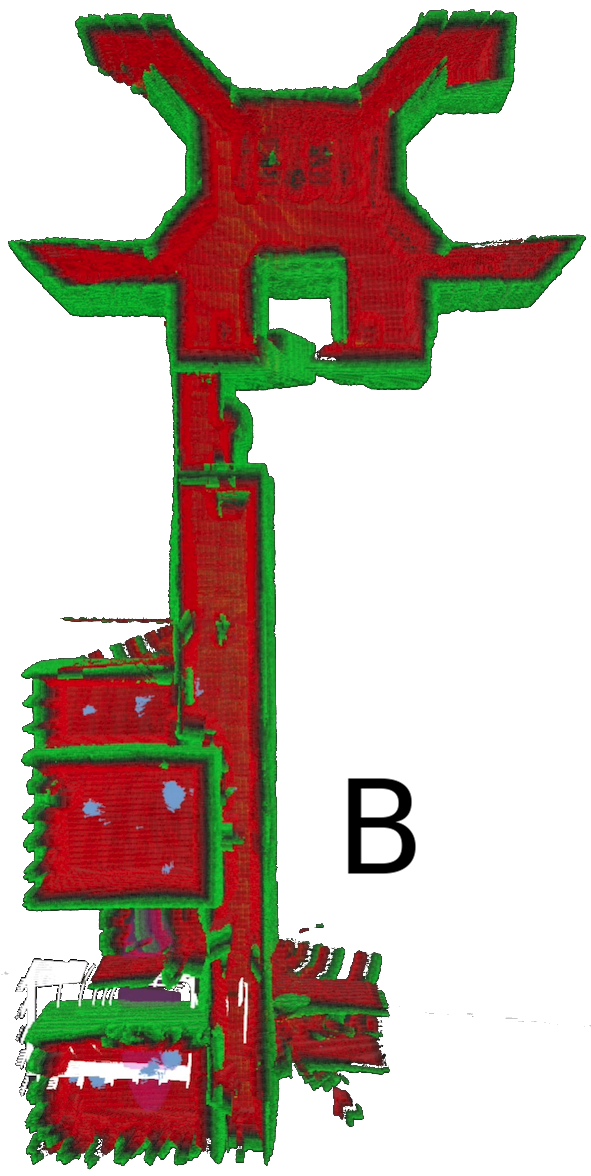}
	\end{subfigure}
	\begin{subfigure}[b]{.32\linewidth}
		\includegraphics[trim={0cm 0cm 0cm 0cm},clip,angle=00,width=0.99\columnwidth]{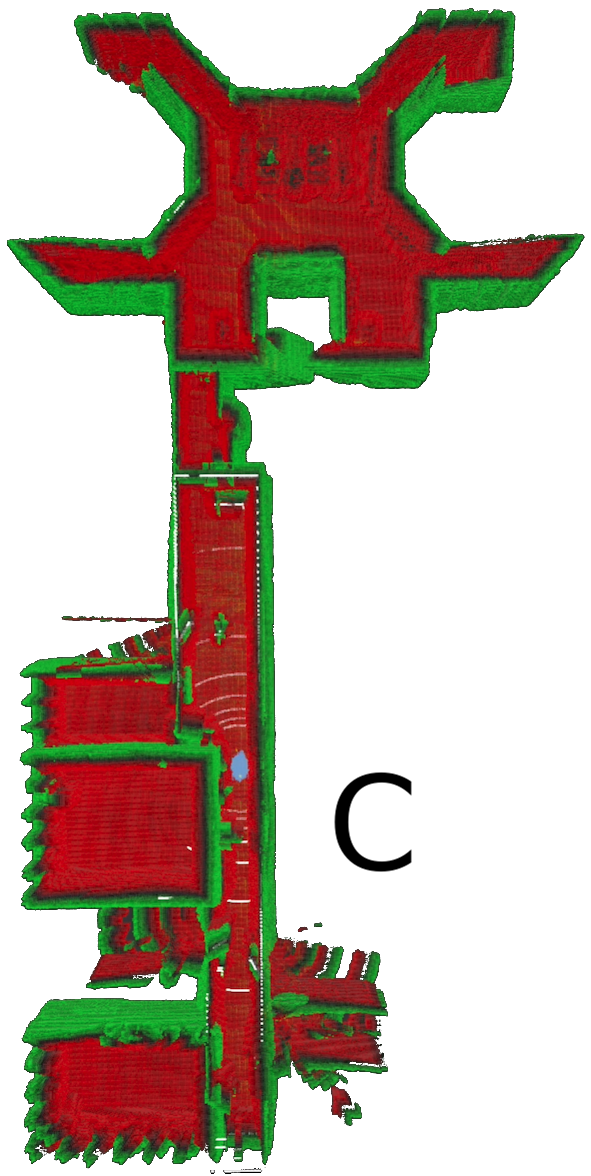}
	\end{subfigure}
	\caption{Our 6D MCL implementation successfully localizes a robot in a large-scale 3D TSDF map, colored in red and green, without initial guess. The multimodal particle distribution is colored in blue and is captured during initialization~(A), after several updates~(B) and near conversion to the real pose~(C).}
	\label{fig:particles}
	\vspace{-0.5cm}
\end{figure}

In this paper, we present an approach that leverages those beneficial properties of TSDF maps and GPUs to achieve a degree of acceleration that allows to use MCL in 6D on embedded hardware.
We provide our implementation in the open-source package \texttt{tsdf\_localization}\footnote{Available under BSD 3-clause license: \url{https://github.com/uos/tsdf_localization}} that is fully integrated with ROS.
In our experiments, we evaluate our software with robots equipped with a LiDAR such as a Velodyne VLP-16 or an Ouster OS0-64, and TSDF maps of various environments.
Our GPU architecture is able to speed up the computation significantly compared to the CPU baseline, while consuming considerably less power.
Moreover, we show that our software is not only suitable for pose tracking but also for global localization of a mobile robot as visualized in \autoref{fig:particles}.

\section{Related Work}

MCL was first introduced by Fox et al.~\cite{fox1999monte}. 
It is an approximation of Markov localization in a 2D grid map and approximates the set of considered states of the robot with a particle filter. 
This significantly increases the performance of the algorithm and allows a global 2D localization with up to 5,000 particles on a CPU.
Using the Kullback–Leibler divergence~\cite{fox2001kld}, the number of particles can be reduced by adapting the number of particles dynamically over time (KLD sampling).
Besides global localization, particle filters are also used in Simultaneous Localization and Mapping (SLAM) algorithms to track the position of the robot and map the environment in 2D using various pose hypotheses~\cite{hahnel2003efficient, grisetti2007improved}. 
Because the initial pose of the robot in the map is known, e.g., Grisetti et al.~\cite{grisetti2007improved} require only 60 particles to track a robot's position.

These approaches concentrate on the localization of robots in 2D maps of planar environments.
For many applications like UAVs in GPS-denied places, localization with six degrees of freedom (DoF) is desirable. 
Perez-Grau et al., for instance, extended MCL to track the position of a flying robot~\cite{perez2017multi}. 
They used landmarks for pose tracking, but did not consider the global localization problem. 
Even in this simplified scenario, they already required 300 particles.
The global localization problem in 3D is addressed in works by Oishi et al. and Hornung et al.~\cite{oishi2013nd, hornung2010humanoid}. 
Their method is not real-time capable and requires between 20,000 and 72,000 particles to localize a robot in a 3D map, which is ten times more compared to the 2D case. 
K{\"u}mmerle et al. also presented an approach for MCL in 3D to localize a robot in an outdoor environment~\cite{kummerle2008monte}. 
They stated that in their scenario at least 100,000 particles are required to achieve basic localization.
All these papers demonstrate the high computational demands of the problem and state that, in comparison with the 2D case, significantly more particles are required.

Due to the parallel nature of the particle filter, in recent years parallel hardware architectures have been used to speed up computations.
Existing methods using such hardware only consider local tracking and can be divided into GPU-based and FPGA-based approaches. 
Examples for GPU-accelerated algorithms are~\cite{fallon2012efficient, kanai2015improvement, dhawale2018fast, rahmangpu}.
In the experiments presented in these works, a fixed number of up to 1,000 particles has been considered, far less than required for realistic applications.
Related methods using FPGAs are presented in ~\cite{bolic2005resampling, liu2014parallel, abd2012proposed, krishna2020source}. 
They consider the general algorithm of the particle filter to solve the bearing-only tracking problem, which is not as complex as 6D MCL. 
Other existing works like~\cite{cho2006real, saha2010design} use particle filters to track objects in camera images.
Such tracking problems are less complex than global localization, hence tracking is possible with only 300 particles.

Based on this review of existing literature, it can be concluded that global MCL in 6D is still unsolved for realistic applications. 
We estimate, that at least 100,000 particles have to be considered when solving the problem for a realistic indoor environment. 
Since mobile robots usually have a limited power budget, solutions have to be found that allow to simulate such high numbers with minimal energy consumption. 

In the case of particle filters applied to the problem of robot localization, GPUs have already achieved a high acceleration of the algorithm and are highly promising to get a significant step closer to the global localization in 6 DoF, where FPGA-based implementations have mostly dealt with less complex scenarios and are yet not as high investigated for this kind of problem as the GPU.

In this work, we provide an implementation for embedded GPU-based architectures to accelerate MCL in 3D TSDF maps to investigate the potential of developing a mobile robot system with 6 DoF.
We provide details on the implementation, including mechanics to ensure efficient memory access, which proves to be important to avoid bottlenecks in the parallel computation.
In the evaluation, we present a use-case, where a mobile robot is localized in a simulated office environment and also analyze the architecture using an established real-world benchmark dataset.
We show, that our implementation is able to solve the global localization problem significantly faster while reducing the power consumption clearly compared to standard CPU-based hardware.

\section{Algorithm}
\label{Sec:algo}

\begin{figure}[b]
	\centering
	\includegraphics[width=0.485\linewidth]{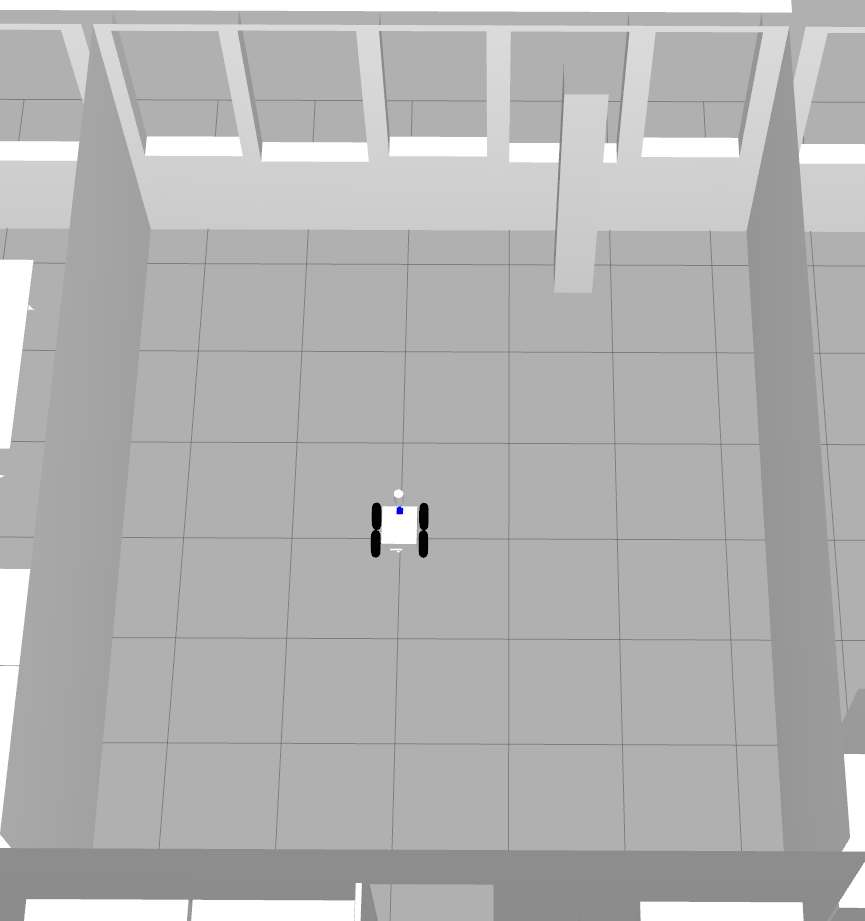}
	\vspace{0.1cm} 
	\includegraphics[width=0.485\linewidth]{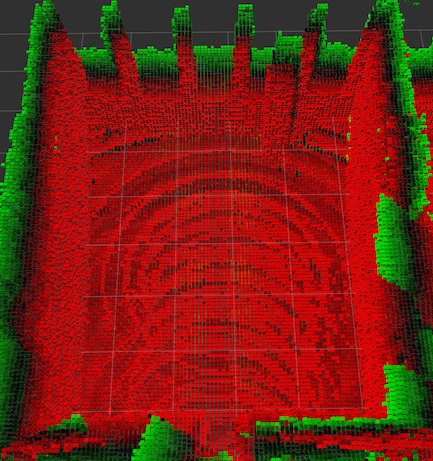}
	\caption{Example of the simulation environment~(left) with the robot model and the resulting TSDF volume~(right). 
	Red cells represent positive values and green cells represent negative values. 
	Darker cells are closer to the surface. 
	Cells exceeding the truncation distance are not rendered.}
	\label{fig:TSDF_example}
\end{figure}

As our algorithm is similar to standard MCL~\cite{fox1999monte}, this section only briefly summarizes that approach, concentrating on the special properties of TSDF maps.
They show several properties that are beneficial for our purposes.
First, the map can be efficiently represented in a sparse 3D array. 
Second, although the representation itself is discrete, for each possible pose, a pseudo-continuous state can be estimated via interpolation.
Secondly, for each hypothesis and a given scan, only a simple lookup is required to obtain the closest distance to the map, which immediately results in an update of the sensor.
Considering these properties, the transfer from standard MCL to 6D MCL is straight forward, although computationally demanding.

To support fast evaluation of the sensor model, a TSDF map is used.
Similar to Akai et al.~\cite{akai20203d}, it is split up into a fine and a coarse-grained representation to save memory while allowing fast access. 
The map only holds parts in the coarse grid, which contain distance values within the truncation distance. 
For all other grid cells, the values are clamped to the truncation value and can be neglected. 
Hence, the presented method scales well to larger environments.
An example of such a map is shown \autoref{fig:TSDF_example}.

In the first step of the algorithm, the particle set is initialized based on the given initial distribution of the robot's state. 
Each particle is represented as described in \autoref{eq:partikel} and \autoref{eq:pose}. 
It holds the robot state $s_t$ in 6D (position and rotation) and a weight $w_t$. 
\begin{equation}
\label{eq:partikel}
p_t^{[n]} = (s_t^{[n]}, w_t^{[n]} )
\end{equation}
\begin{equation}
\label{eq:pose}
s_t^{[n]} = (x_t^{[n]}, y_t^{[n]}, z_t^{[n]}, \phi_t^{[n]}, \psi_t^{[n]}, \theta_t^{[n]})
\end{equation}
If a pose estimation is available, the particles are distributed with predefined variance around the initial state estimation. 
Without an initial pose estimate, the particles are distributed uniformly in free space. 
After initialization, three steps of the algorithm are repeated to iteratively improve the pose estimation of the robot. 
The first step is the motion update that applies the motion model to each particle.

After the propagation of the robot's state, the sensor update is computed, where the likelihood for every particle is calculated using the sensor model of the robot. 
In this step, a virtual sensor measurement for every particle is generated and compared to the map. 
The weight of a particle increases with the match of the observation to the environment. 
In this work, the endpoint model is used to estimate the weights, where the distance of the scan points to the surface of the environment is used as measure of the match.
Because of the structure of the TSDF representation, this can be calculated directly using \autoref{eq:likeli_df}. 
The weight for every particle depends on the considered pose $x_t$ in the map $m$.
$m{(x_t, z_t)}$ represents the signed distance of the considered scan point $z_t$ from the perspective of the particle to the next obstacle in the map, which can be obtained directly by a look-up in the TSDF volume. All entries of the map can be calculated before the start of the scan procedure, where the inefficient calculation of the exponential function can be prevented. However, using a laser scanner the sensor update is still computation expensive, because a high number of scan points must be evaluated for every single particle. 
\begin{equation}
\label{eq:likeli_df}
p_{\text{hit}}(z_t|x_t^{[n]}, m) \approx \frac{1}{\sqrt{2 \cdot \pi \cdot \sigma^2}} \exp{(-\frac{1}{2} \cdot \frac{m{(x_t^{[n]}, z_t)}^2}{\sigma^2})}
\end{equation}
The pose can then be estimated by the weighting average of the particles or by using the particle with the highest weight as the current pose of the robot.
The last step is to draw a new set of particles. 
This is done using the state distribution represented by the current particle set. 
Based on the calculated weights, a new set of particles is chosen from the previous one in the resampling step. 
Of all the steps of the algorithm, the sensor update is by far the most computationally intensive and its acceleration is therefore the focus of this work.

\section{System Architecture}

\begin{figure}
	\centering
	\includegraphics[width=\linewidth]{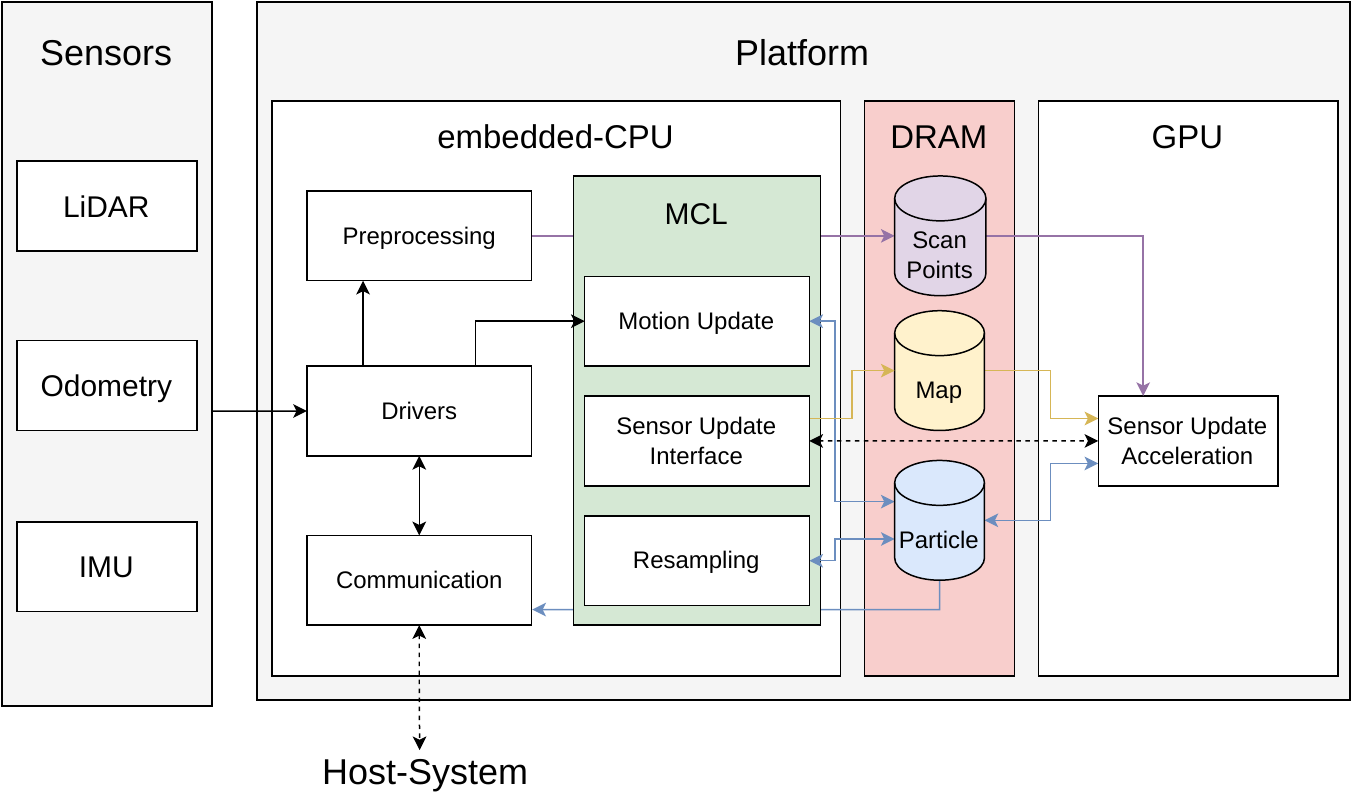}
	\caption{Visualization of the system architecture for the GPU.} \label{Fig:system}
\end{figure}

In this section we describe the basic system architecture for the CPU and GPU implementation of the algorithm, where the CPU-based implementation serves as the baseline to evaluate the acceleration based on the embedded GPU.
To ensure comparability, all systems have the same structure and are able to receive and send data via the same interface, which allows to evaluate them on the same datasets.
Moreover, the GPU-based implementation can be considered as an extension of the CPU-based implementation.  
An overview of the system architecture is shown in \autoref{Fig:system}.
In both cases embedded platforms with CPU and GPU are used.
As described in \autoref{Sec:algo}, the sensor update is the most computationally expensive part of the algorithm. 
Therefore, this step is executed on the acceleration hardware, while the motion update and the resampling remain on the embedded CPU.
The systems are able to operate without the need for an external computer, but can also communicate with a host using ROS to receive sensor data, visualize the results or replay pre-recorded data via bag files.
Odometry estimation for the motion model update is provided by dead reckoning or IMU. 
3D point clouds are received from a laser scanner to evaluate the current particles in the TSDF map. 
On GPU-based platforms, an interface process is running on the CPU to handle the access to the GPU.

\section{Implementation}

The provided implementations to accelerate the evaluation of the particles on all hardware platforms are detailed in this section. 
The main algorithm of the sensor update consists of two consecutive steps.
First, the sampled poses $p_t^{[n]}$ must be evaluated to compute the weights $w_t^{[n]}$ of the particles. 
Afterwards, the weights are used to determine the weighted sum of the considered poses to get a pose estimation of the robot system. 
These steps have to be implemented differently with respect to the underlying hardware architecture.  

\begin{figure}
	\centering
	\includegraphics[width=\linewidth]{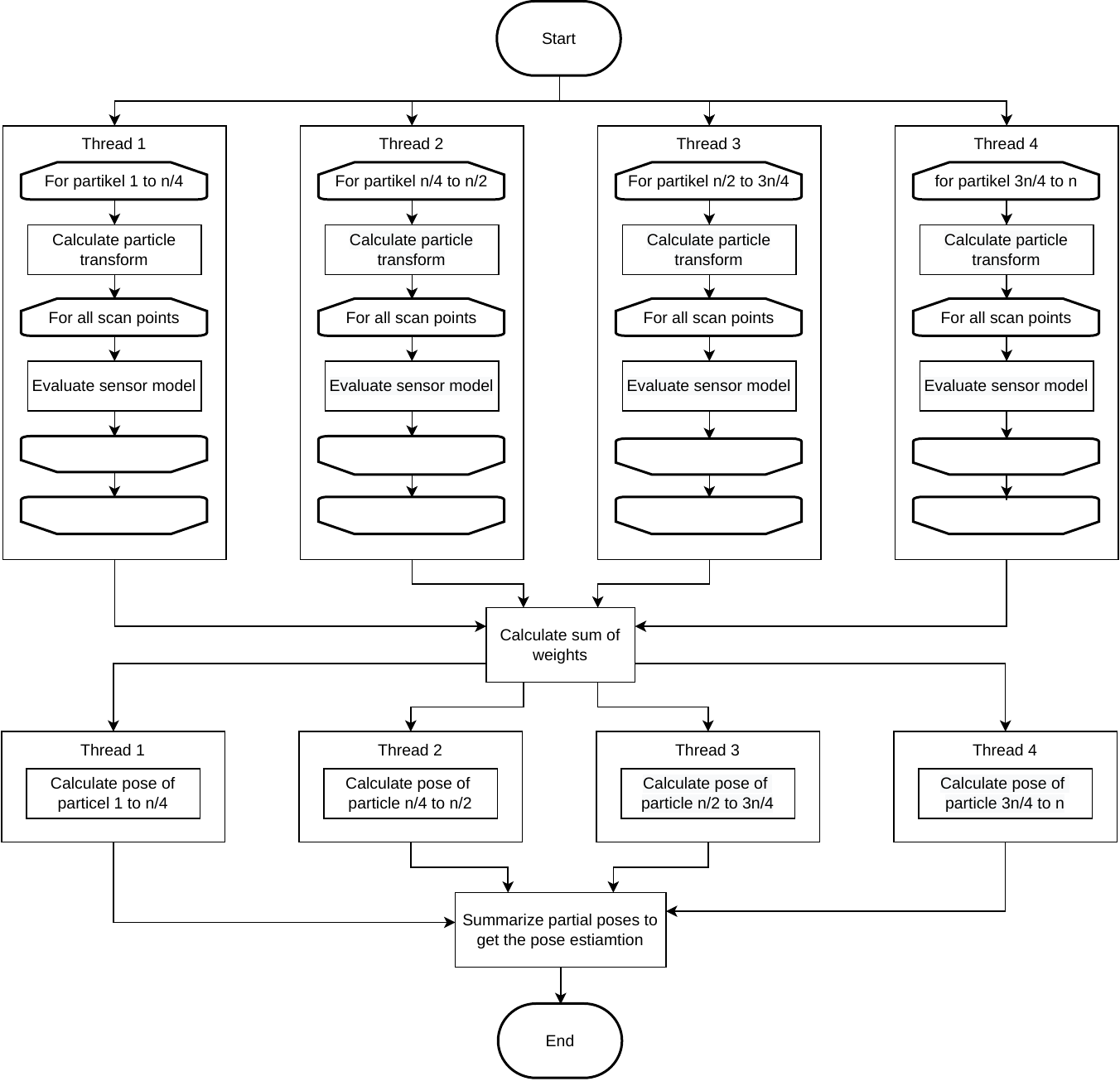}
	\caption{Program flow of the CPU-based implementation with four available threads. The particles are split up into groups according to the number of available threads.} \label{Fig:cpu_imp}
\end{figure}

\subsection{CPU Implementation}
\label{Sec:cpu_imp}

For comparability in terms of power consumption and performance, all available threads of the CPU-based system have to be used to achieve the maximum utilization of the hardware. 
The program flow of the implementation is shown in \autoref{Fig:cpu_imp}. 
The particles are split up into groups according to the number of available threads in the system. 
Every group of particles is processed in parallel by an assigned single thread. 
After evaluation of all particles, the threads are synchronized to sum up the particle weights, which is required for normalization. 
The calculation of the pose estimation is also split up into the available threads similar to the evaluation procedure. 

\begin{figure}
	\centering
	\includegraphics[width=\linewidth]{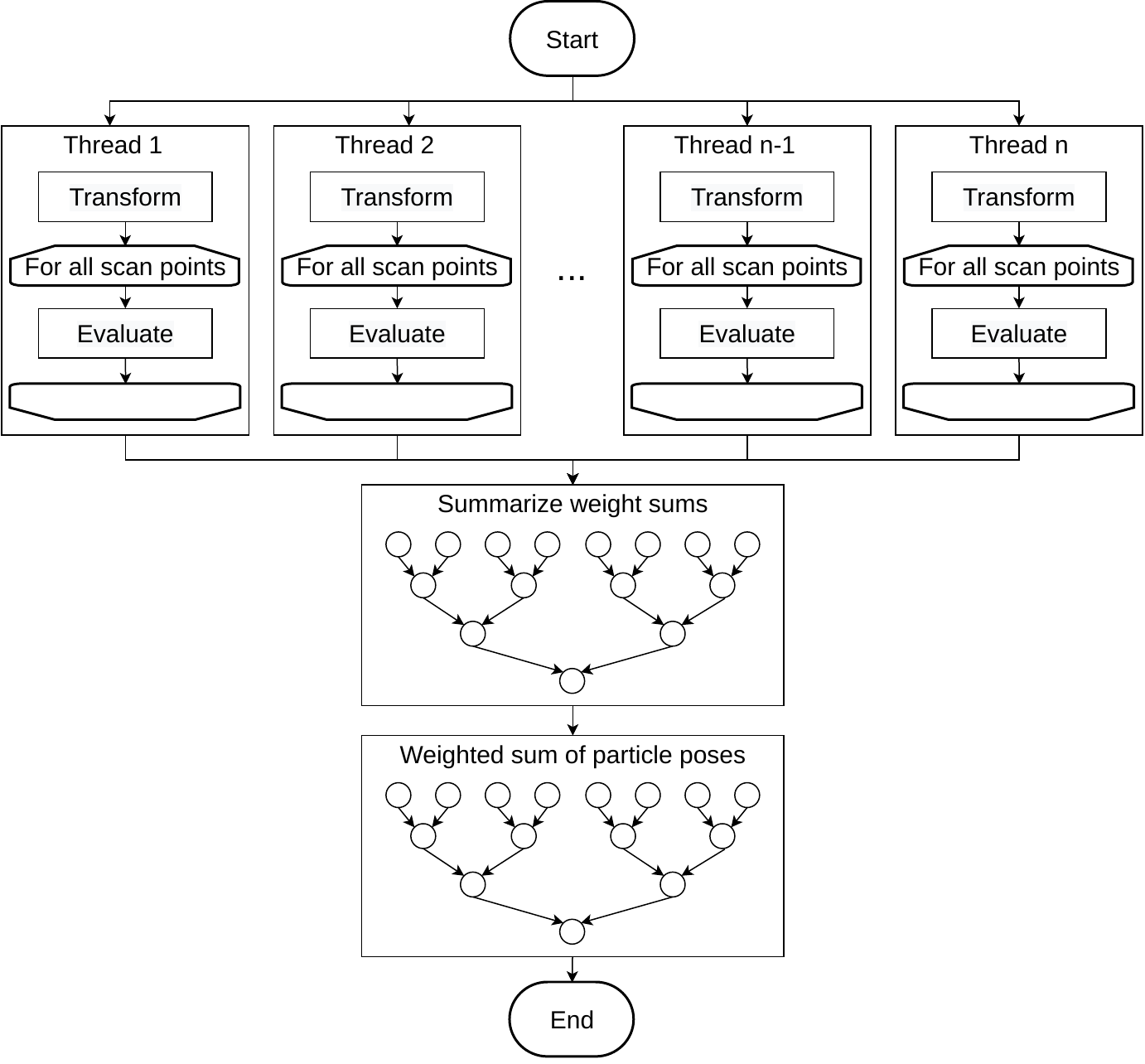}
	\caption{Program flow of the GPU-based implementation. Each particle is processed by single thread.} \label{Fig:gpu_imp_per_part}
\end{figure}

\begin{figure}
	\centering
	\includegraphics[width=\linewidth]{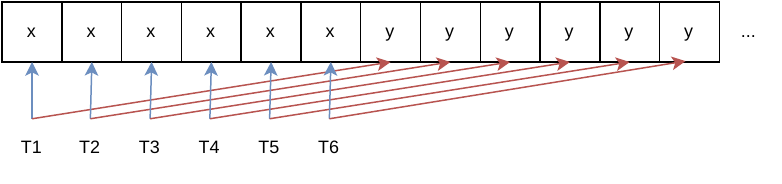}
	\caption{Visualization of the access pattern for the particles and the scan points per thread on the GPU. Every letter stands for one coordinate of the particles or the scan point.} \label{Fig:access_pattern}
\end{figure}

\begin{figure}
	\centering
	\includegraphics[width=0.7\linewidth]{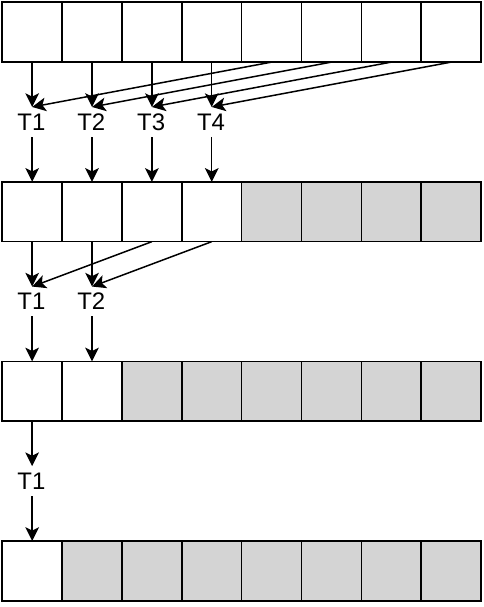}
	\caption{Simplified example of the reduction pattern used to merge the partial results on the GPU. Each thread summarizes two data in every iteration until only one final data point is left. The gray fields mark data points no longer considered by the algorithm.} \label{Fig:reduction}
\end{figure}

\subsection{GPU Implementation}

For maximum throughput on GPUs, many independent threads have to be scheduled to hide latency inside the system. 
The idea is to adapt the CPU-based implementation from the previous section by processing every particle in a single GPU thread. 
This is visualized in \autoref{Fig:gpu_imp_per_part}. 
Despite the property, that all particles can be evaluated independently, one major bottleneck to be dealt with is the high amount of memory access on the particles, scan points and map data.
Although the map access is highly unstructured and strongly depends on the distribution of the particle cloud, the access to the particles and scan points in the algorithm is structured.
Hence, the memory efficiency can be optimized by ordering the data according to the threads in the multiprocessors, as visualized in \autoref{Fig:access_pattern}.
Although this leads to higher parallelism, it also increases the synchronization overhead for computing the particle weights and the pose. 
This is handled by implementing a GPU-based reduction pattern, which exploits the memory hierarchy of the GPU. 
Depending on the number of used particles in MCL, the reduction kernel needs to be called several times to achieve synchronization of all threads. 
This is because of the processing structure in the GPU, which allows synchronization between threads only within a thread block of limited size. 
So every iteration of the reduction pattern summarizes the thread blocks hierarchically to exploit parallelism, as shown in \autoref{Fig:reduction}.
In every iteration, each thread sums up two data points. 
Thus, in every iteration the number of data points and active threads is halved until the final result is reached.
Similar to the particle ordering, the threads access and store data in the same order as they are executed on the multiprocessors of the GPU, resulting in a better utilization of the memory bandwidth.

\begin{figure*}
	\centering
	\includegraphics[width=0.32\textwidth]{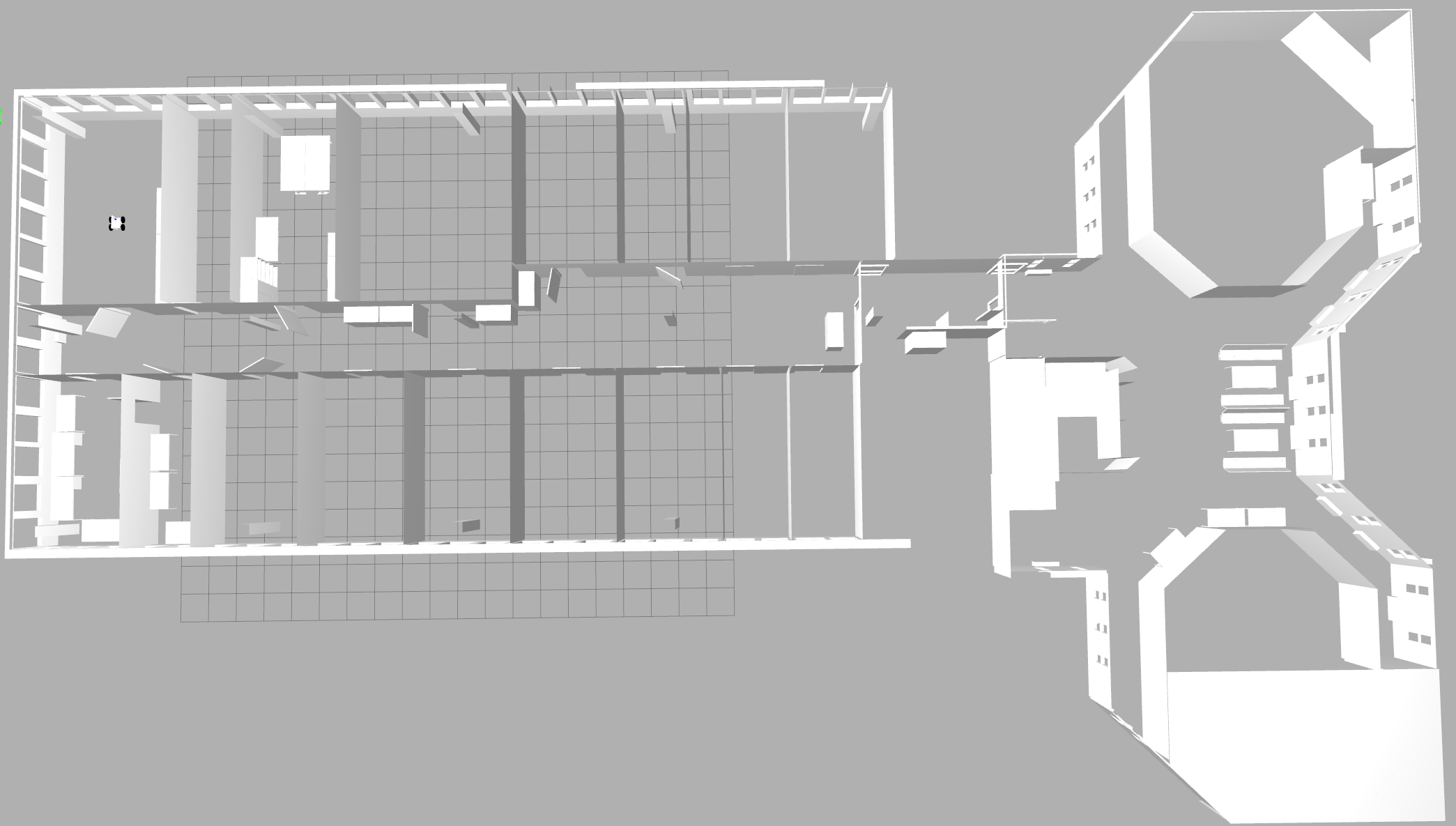}
	\includegraphics[width=0.32\textwidth]{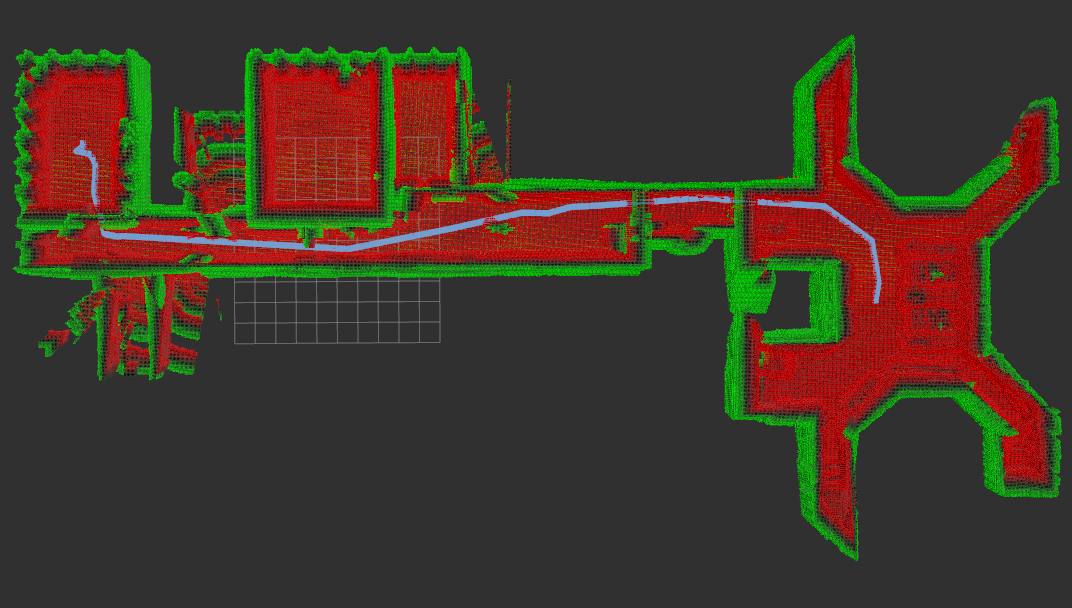}
	\includegraphics[width=0.32\textwidth]{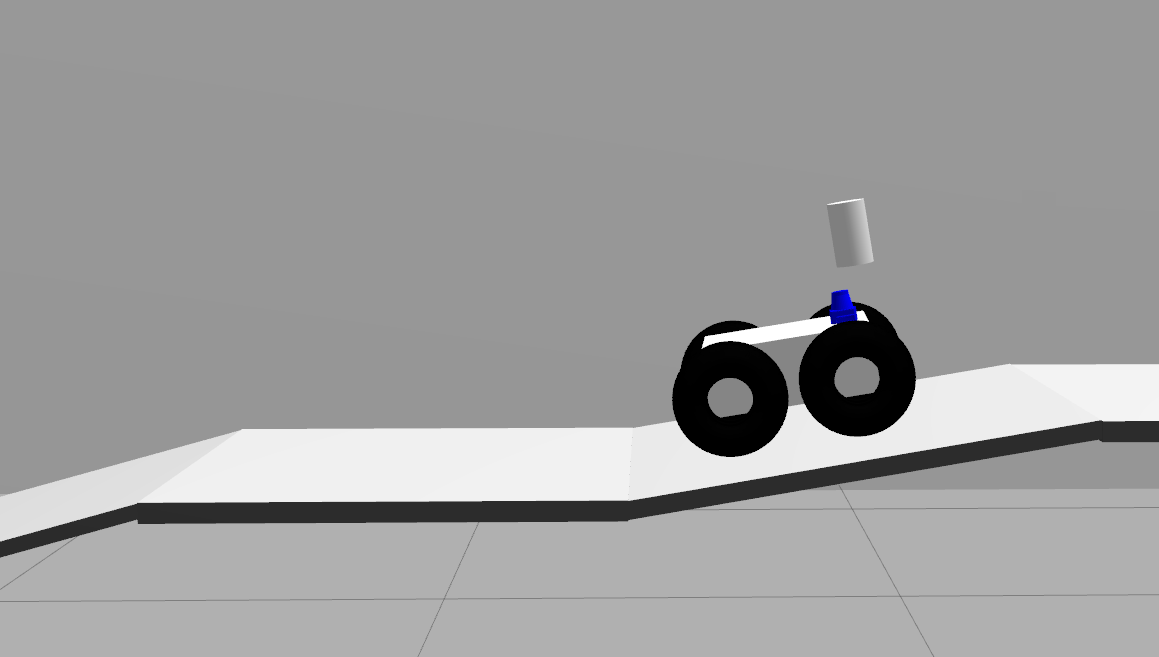}
	\caption{Top down view of the simulation environment~(left), the corresponding TSDF map~(middle) and the ramp scenario~(right). The traveled path of the robot is marked blue.}
	\label{fig:env}
\end{figure*}

\begin{figure*}
	\centering
\includegraphics[width=0.32\textwidth]{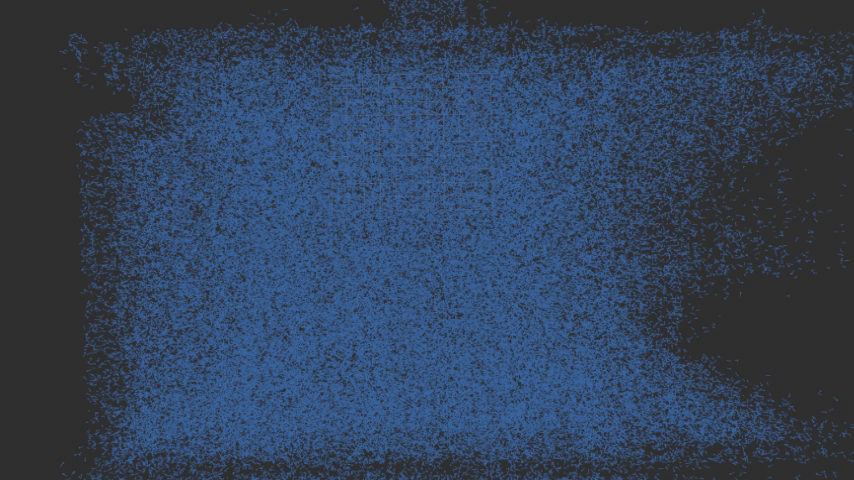}
\includegraphics[width=0.32\textwidth]{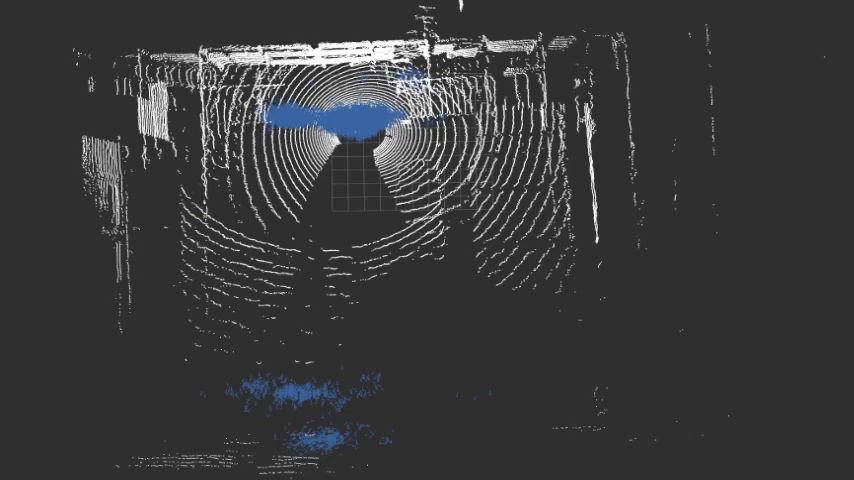}
\includegraphics[width=0.32\textwidth]{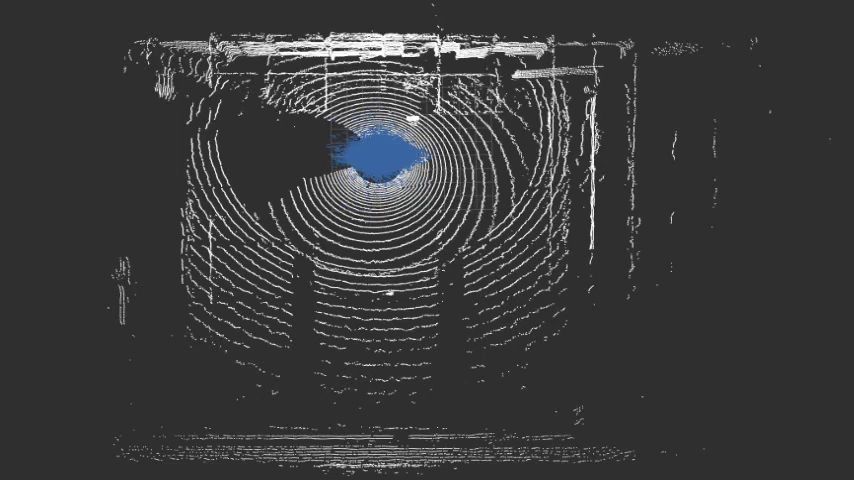}
\caption{Top down view of the particles during the initialization~(left), after several updates~(middle) and near conversion to the real pose~(right) while processing HILTI's Drone-Test arena dataset.}
\label{fig:hilti_particles}
\end{figure*}

\section{Evaluation}

To evaluate the implementations for the CPU and the GPU we used pre-recorded datasets.
First, we used an office floor, shown in \autoref{fig:env}, to test our implementation in a common GPS-free indoor environment.
It covers an area of $50 m \times 20 m$.
In addition, a robot equipped with a virtual Velodyne VLP-16 LiDAR was simulated in this environment using Gazebo.
In the original Gazebo environment a safely driving robots moves on the $xy$-plane only.
To demonstrate the localization of a robot moving in 6D state space, we placed ramps into the map as shown in the left part of \autoref{fig:env}.
Using a simulation environment lets us investigate an arbitrary number of motion trajectories within the same map where each motion trajectory is provided with a ground truth localization.

Second, we used datasets of the HILTI SLAM Challenge \cite{hilti} to especially evaluate the GPU-based implementation to its capability to globally localize a robot in real-world indoor scenarios.
We focused on the "Drone-Testing arena" sequence because it contains sensor data acquired from an Ouster OS0-64 and, most importantly, provides a 6-DoF trajectory of the sensor as ground truth.
We first use the ground truth as a perfect odometry estimation for the motion update to avoid false odometry estimations affecting the global localization.
From there on, we could systematically increase the level of applied noise which allows us to make accurate predictions about how much odometry estimation error is allowed in order for our system to still produce reliable global localization results. 

For both the simulated and the real-world experiments we generated the required TSDF maps using HATSDF-SLAM \cite{eisoldt2021hatsdf}.
Then, the sensor data and the respective TSDF map is passed into our software via the provided ROS interface to ensure comparability.
During the following experiments we tested our implementation for a large number of trajectories, observing whether and how the algorithm converges, and additionally measuring the localization accuracy, performance, and power consumption.
To ensure the generalization of our implementation, the evaluation experiments use trajectories different then those used to create the map.

\begin{figure*}
	\centering
	\includegraphics[width=0.49\textwidth]{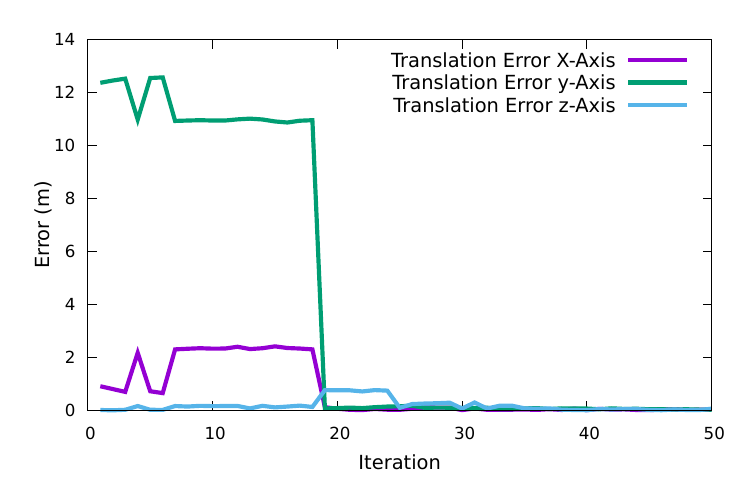}
	\includegraphics[width=0.49\textwidth]{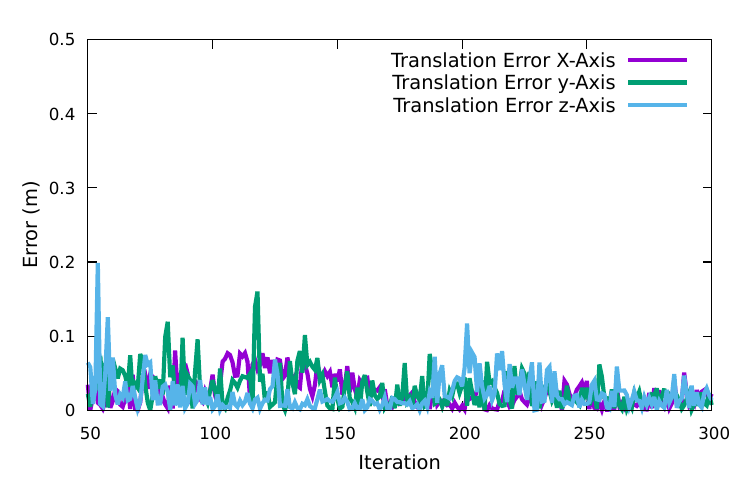}
	\caption{Translation error for every axis during global pose estimation~(left) and the following pose tracking~(right) of the using the simulation dataset.}
	\label{fig:trans_err_sim}
\end{figure*}

\begin{figure*}
	\centering
	\includegraphics[width=0.49\textwidth]{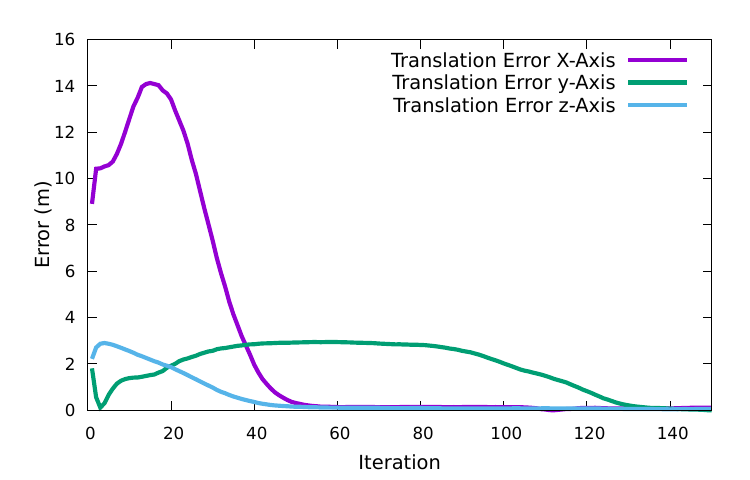}
	\includegraphics[width=0.49\textwidth]{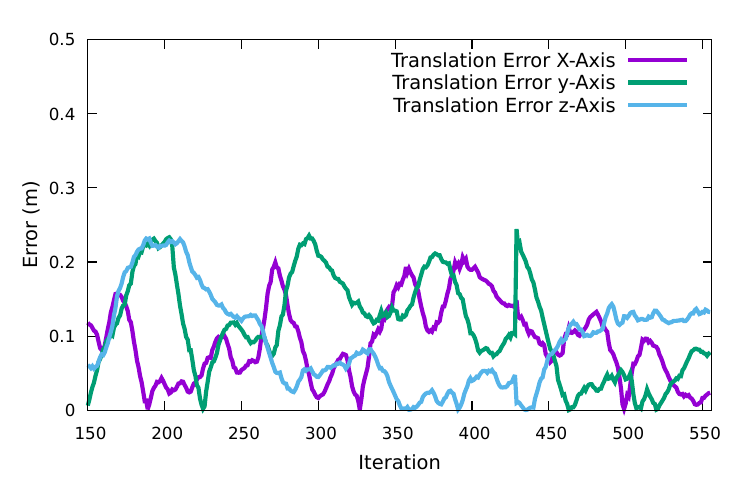}
	\caption{Translation error for every axis during global pose estimation~(left) and the following pose tracking~(right) using HILTI's Drone-Test arena dataset.}
	\label{fig:trans_err_hilti}
\end{figure*}

\subsection{Qualitative Evaluation}

First, we test our implementation in the simulated office environment.
\autoref{fig:particles} shows an example experiment.
Initially, the particles were uniformly sampled from the free space and with six degrees of freedom.
After a few iterations of MCL, the particles are distributed around several probable poses, reflecting symmetries and the similarities of the rooms.
Finally, the particle distribution converges around the correct pose of the robot.
White dots (best seen in the PDF file) visualize the transformed point cloud of the laser scanner, which, once it matches the map, indicates successful localization.
The reconstructed trajectory in this scenario overlaps with the ground truth, as visualized in \autoref{fig:env} (middle).
For this run, the translation errors for every axis is plotted in \autoref{fig:trans_err_sim}.
Repeating the experiment for 11 different trajectories shows that approximately 50,000 particles are needed to achieve robust localization of the robot in this particular environment.
Furthermore, both the CPU and GPU implementation achieves a localization accuracy in all simulation experiments within the order of magnitude of the fine map resolution (about 6\,cm).

Besides the simulated scenarios, we also evaluated the quality of the global localization in a real-world environment of the HILTI datasets.
Similar to the simulated environment, the particles are initially distributed uniformly in the complete free space of the environment.
For the Drone-Testing arena, the distribution of the particles for different iterations can be seen in \autoref{fig:hilti_particles}.
Similar to the simulated scenario, several particle clusters are formed at the beginning, but converge against the true pose of the robot.
We found that increasing the noise of both the velocity and angular velocity components of the motion estimate to $\sigma=0.1$ does not significantly affect the convergence speed.
When setting the noise $\sigma$ to 0.1, 100,000 particles are needed to achieve a reliable global localization in this scenario.
The convergence of the estimated pose can also be seen in \autoref{fig:trans_err_hilti}, where the translation error for every axis is plotted for the Drone-Testing arena sequence.
The results show that the localization error increases at the beginning.
This is because of the multiple particle clusters, which lead to a wrong average pose estimation.
Then, the pose error decreases significantly as the individual clusters disappear.
Overall, the results on the HILTI datasets show that our software is able to perform global localization even in real-world, feature-less environments that contain many symmetries, while moving freely in 6DoF space.

\subsection{Performance}

\begin{figure}
	\centering
	\includegraphics[width=\linewidth]{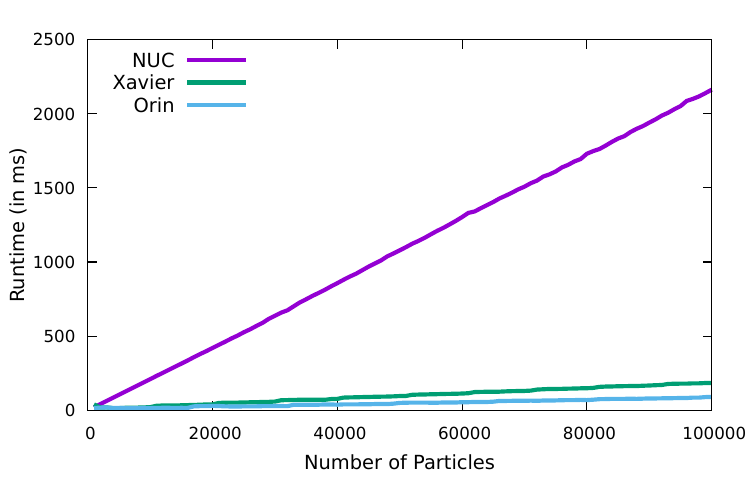}
	\caption{Comparison of the measured run times with an increasing number of particles during a normal distribution.} \label{fig:runtime_compare}
\end{figure}

In case of the simulated robot, the laser scanner can provide point clouds with a frequency of 20 Hz.
Therefore, the implementations have to compute an iteration of the sensor update in less than 50\,ms to achieve real-time performance, while the Ouster laser scanner provides scan points with 10 Hz. Hence, the sensor update must be computed within less than 100\,ms to achieve real-time performance in the HILTI datasets. 
To evaluate the performance, an Intel NUC (NUC6i7KYK, Core i7-6770HQ) served as CPU baseline.
For the GPU implementation, we used a Jetson AGX Xavier and a Jetson AGX Orin to accelerate the algorithm and to show the scalability of the implementation.
To compare the performance of the implementations, the run time of the sensor update was measured for normally distributed particle clouds with an increasing number of particles. 
The measured run times are shown in \autoref{fig:runtime_compare}. 
As expected, the run time of all implementations scales linearly with the number of particles.

In addition, the run times of all implementations were measured during the localization on the traveled path in the simulated environment. 
\autoref{tab:50000energy} shows that although the goal of being faster than the maximum scanning frequency of 20 Hz was not achieved, both accelerations based on the GPU are able to speed up the sensor update significantly.

\begin{table}
	\centering
	\caption{Performance and power consumption of the implementations on the CPU (NUC) and GPU (AGX Xavier and Orin) in the simulated environment.}
	\label{tab:50000energy}
	\begin{tabular}{lccc}
		\toprule
		& NUC & AGX Xavier & AGX Orin \\
		\midrule  
		Run time [ms] & 1527 & 155 & 51 \\
		Power [W] & 72.2 & 34.64 & 41.2 \\
		Energy [J/Particle Cloud] & 110.25 & 5.37 & 2.1 \\
		\bottomrule
	\end{tabular}
\end{table}

\subsection{CUDA Metrics of the GPU-based Implementation}

The GPU-based implementation has been developed using CUDA. So various metrics can be achieved to analyze the efficiency of the implementation on the hardware. They are estimated using a prerecorded normal distributed particle cloud with the tool nvprof. The most important metrices for the particle cloud are depicted in Tab. \ref{tab:cuda_stats}. As can be seen, the implementation achieves a high occupancy of the GPU. The remaining resources of the GPU cannot be occupied, which occurs because of the look-ups in TSDF map. 
The random look-up of a TSDF value can lead to one or two memory accesses to the DRAM, depending on the occupancy of the requested grid cells. So that threads finish their work at different times, leading to inactive resources.

The random distribution of the particles in the environment of the robot has also an impact on the performance and efficiency of the memory access. The look-up of the TSDF values reduces the cache hits significantly. Although some look-ups share the same coarse grid, the most memory access for the transformed scan points are widely distributed in the environment resulting only in a few shared memory access on the fine map between the scan points. The widely and randomly distributed particles also lead to unaligned memory access in the TSDF map between neighboring threads. This reduces the efficiency of the read accesses, because only a few parts of the read data blocks read from DRAM can be used. Furthermore, the cache hits decreases as the particle cloud is distributed more widely in the environment. When the particles are initialized globally the L1 cache hits decreases to 31.53 \% and the L2 cache hits achieved a value of 11.47 \%
However, the writing accesses into DRAM can be performed with the maximum efficiency, because all threads write the weights of the particles aligned back into the global memory.         
All in all it can be concluded, that the memory accesses to look up the TSDF value in the map of the environment are the major bottleneck for the implementation on the GPU, because the underlying hardware architecture does not fit the kind of reading accesses. 

\begin{table}
	\centering
	\caption{Considered CUDA metrics for the GPU-based implementation for evaluating a locally initialized cloud of 50,000 particles.}
	\label{tab:cuda_stats}
	\begin{tabular}{lc}
		\toprule
		Metric & Value \\
		\midrule
		Achieved Occupancy & 74,163 \% \\
		L1/TEX Cache Hits & 39,64 \% \\
		L2 Cache Hits (read)& 39,63 \% \\
		Global Memory Efficiency (read) & 46,43 \% \\
		Global Memory Efficiency (write) & 100,00 \% \\
		\bottomrule
	\end{tabular}
\end{table}

\subsection{Energy Efficiency}

For localization on a mobile system, power consumption is a crucial factor.
We measured the power consumption via a shunt directly on the boards.
The results are shown in \autoref{tab:50000energy}.
With the GPU implementations, a significant reduction in both power and energy has been achieved. Using our implementation on the AGX Orin resulted in a more than 50 times decrease in Energy per Particle Cloud compared to the CPU implementation.

\section{Conclusion and Future Work} 

Because of the significant acceleration and the high energy efficiency the GPU implementation compared to the CPU-based baseline, a decisive step has been made in the direction of a real-time capable global MCL for mobile robots. 
Our GPU implementation outperforms a CPU implementation by factor of 30, while increasing the energy efficiency significantly by a factor of more than 50.
The next step to improve performance and energy efficiency is to build heterogeneous architectures to extend the GPU-based architecture with specialized units to process particles. 
Furthermore, to evaluate the quality of the invented algorithms in realistic scenarios, new benchmark datasets are required that allow to evaluate many trajectories in challenging environments.
To tackle this problem, we plan to provide a repository of reference TSDF maps with many different trajectories captured with a laser tracking system to provide a benchmarking environment for the development of such algorithms similar to the established KITTY~\cite{Fritsch2013ITSC} and Hilti~\cite{zhang2022hilti} datasets for SLAM.

\bibliographystyle{IEEEtran}
\bibliography{papers.bib}  

\balance

\end{document}